# A REINFORCEMENT LEARNING BASED DECISION SUPPORT SYSTEM IN TEXTILE MANUFACTURING PROCESS*

ZHENGLEI HE [†] & KIM-PHUC TRAN & SEBASTIEN THOMASSEY & XIANYI ZENG

*ENSAIT, GEMTEX – Laboratoire de Génie et Matériaux Textiles, F-59000 Lille, France*

CHANGHAI YI

*Wuhan Textile University, 1st, Av Yangguang, 430200, Wuhan, China*

This paper introduced a reinforcement learning based decision support system in textile manufacturing process. A solution optimization problem of color fading ozonation is discussed and set up as a Markov Decision Process (MDP) in terms of tuple {S, A, P, R}. Q-learning is used to train an agent in the interaction with the setup environment by accumulating the reward R. According to the application result, it is found that the proposed MDP model has well expressed the optimization problem of textile manufacturing process discussed in this paper, therefore the use of reinforcement learning to support decision making in this sector is conducted and proven that is applicable with promising prospects.

## 1. Introduction

Competition in the textile industry is increasingly influencing the firms, especially in the field of manufacturing. While a large number of processes involved in textile manufacturing with the interaction of a wide range of variables in each single process, which obviously illustrates its difference from other areas and hence it hardly can be promoted by only taking reference in respect of the engineering technologies but taking optimization into consideration.

In terms of manufacturing optimization, producers have to select the most effective solutions from various combinations of different processes with the proper selection of materials and parameters. Decision making plays a key role in this issue. Due to the fact that the relationship between the variables and product properties is not well established, the decision maker is unaware of the probabilities of future states, which turns out that the decision making mostly is under uncertainty or risk [1].

---

* This work is partially supported by China Scholarship Council (CSC, Project NO. 201708420166).
[†] Correspondent author : zhenglei.he@ensait.fr

Decision support system can help decision makers by taking advantage of models, data and knowledge on the basis of artificial intelligence [2]. The application of it has been conducted in certain textile processes using machine learning [3], genetic algorithm [4] and fuzzy technology [5] etc. In this work, however, we proposed a novel decision support system based on reinforcement learning (RL) to deal with the production optimization problems in textile manufacturing.

RL is an algorithm learning to take actions in a dynamic environment from a particular state to transit to a new state by acquiring immediate rewards and maximize the accumulative reward over time from the interaction with environment [6]. The dominant problem RL tackle with is Markov Decision Process (MDP) in the formula of a tuple : {*S, A, P, R*}, where *S* is a set of environment states, *A* is a set of actions, *P* is a transition function, *R* is a set of reward or losses.

In order to present the applicability of the developed RL based decision support system, a case study is given on the cost optimization of a color finishing process in textile manufacturing. Following sections also include the description of problem formulation and framework construction before the discussion of case study application.

## 2. Proposed framework for optimizing textile manufacturing process

### 2.1. *Problem formulation*

Decision making in textile process involves a wide range of variables, finding a solution with the best selected parameters in a process is always time consuming as the traditional methods are dependent on trial and error on the basis of experience. In recent years, researches regarding predictive models based on various regression approaches or machine learning algorithms, such as support vector regression and artificial neural network have made a tremendous difference in this situation. The relationship of variables is not completely unclear in certain cases. In other words, decision making in textile manufacturing can be conducted under certainty or risk when the process is well modelled with excellent predictive power. While due to the large number of variables, the future state of nature or the probabilities associated with all feasible future states of nature in terms of the combination of parameters in different solutions can be immense.

In the view of RL, present problem can be tackled with training an agent to traverse and explore the state space as a MDP. All the combinations of the targeted parameters are elements composed to the state space, agent learns by taking actions of going up (+), going down (-) or staying (0) from a given state in the environment depending on cumulating the state-action value feedback from the environment. In which the feedback from the environment is based on a reward function in accordance with the objective function in a specific optimization problem.

### 2.2. Q-learning algorithm

The Q-learning is one of the most widely used value-based RL algorithms. It is a learning approach by estimating the sum of rewards $r$ for each state $s$ when a particular policy $\pi$ (e.g. ε-greedy policy) is being performed. It uses a tabular representation of the value $Q^\pi(s, a)$ to perform an action $a$ in a state $s$ and iteratively update the *Q-value* according to the following rule:

$$Q^\pi(s,a) \leftarrow Q^\pi(s,a) + \alpha[r + \gamma max_{a'} Q^\pi(s',a') - Q^\pi(s,a)]$$

where $s$ and $a$ are the current state and action respectively, while $s'$ is the state achieved when executing $a'$ in the set of $S$ and $A$ in any given MDP tuples of $\{S, A, P, R\}$. $\alpha \in [0, 1]$ is the learning rate, $r$ is the immediate reward, $\gamma \in [0, 1]$ is the discount factor. In order to balance the exploration and exploitation, ε-greedy is used to instruct the action selection between random unexplored actions and ongoing learned policy $\pi$.

### 2.3. Mapping present problem onto RL

We begin with the mapping of present problem onto RL problem by setting up it to a MDP tuple of $\{S, A, P, R\}$.

**State**: As mentioned, in the optimization problem of textile manufacturing process, we may have a large number of solutions with different combination of parameters $S\ (p_1, p_2...p_n)$, this is defined as the state vector in this study for the convenience of agent to search the optimal solution (state) for certain targets.

**Action**: the setting of parameters in textile manufacturing, usually, is distributed discretely in a limited range (e.g. temperature from 20 to 100℃ by 10 increments in each adjustment). For each parameter in the state $S\ (p_1, p_2...p_n)$, the actions in the action vector are defined as hold (0), or increase (+) / decrease (-) of the regular unit increment of the corresponding parameters, respectively.

Therefore, the actions in action space could be $3^n$ when there are n parameters considered in the decision making.

***Transition function*:** the transition function is about the probabilities of agent transit from a state *s* to next state *s'* when an action is conducted. It is 1 (100%) for most states in the environment, while a specific type of states should not be accessible in the given state space. Taking pH as an example, it is known that pH is only possible to be set from 1 to 14, so when a state with $p_i$ out of the normal range, transition probably would be 0.

***Reward function:*** when agent performs an action in a state to transit to a new state, a reward signal would be received to renew the Q value of this state-action for future consideration of action selection. Therefore, an evaluation of the state is needed. Depending on our previous work [7], a random forest regression predictive model is found that powerful enough and applicable in this case, which has been used in this RL framework to evaluate the states by mapping the state to the objective function $f(p_1, p_2...p_n)$. In order to optimize the objective function, the reward function can be the difference between the value of objective function: $r_i = f_i(p_1, p_2...p_n) - f_{i+1}(p_1, p_2...p_n)$.

## 3. Application to quality optimization of a textile finishing process

Color fading is a significant finishing process for specific textile products such as denim to obtain fashion style, but it conventionally was achieved by chemical methods which have a high cost and water consumption, as well as a heavy burden on the environment. Instead, ozone treatment is an advanced finishing process employing ozone gas to achieve color faded denim without a water bath and consequently it reduces less environmental issues. We have developed random forest models to map the process parameters of water-content (*C*), temperature (*T*), *pH* and treating time (*t*) to the color properties in terms of the values of k/s, L, a and b of faded textile. The predictive error of it has been tested and the result shows that it is less than 1.5%. As the fact that the color properties are the most important quality of the treated product, we optimize the quality of this process by minimizing the gap (as shown below) between predicted color of the solutions and targeted color properties (*k/s, L, a, b*), and find the closest solution as the final result.

Objective function:

$$f(C,T,pH,t) = \sqrt{(k/s - k/s')^2 + (L - L')^2 + (a - a')^2 + (b - b')^2}$$

where the *k/s, L, a* and *b* are predicted by inputting the *C, T, pH* and *t* of a solution to random forest model. *k/s', L', a'* and *b'* are the color properties of targeted products.

### 3.1. *Experimental*

#### 3.1.1. *MDP setup*

State space in this case is composed by the solutions with four parameters $(C, T, pH, t)$. And the units of adjustment of them usually are 50, 10, 1 and 1 respectively in the application in the range of [0, 150], [0,100], [1, 14] and [1, 60] respectively.

So it is clear that the action could be any combination of elements from the following sets: [+50, 0, -50], [+10, 0, -10], [+1, 0, -1] and [+1, 0, -1]. The total number of action space should be $3^4 = 81$.

Transition probability is 1 for the states in the given range of state space above, but 0 for the states out of it. The function is given below:

$$T = \begin{cases} 1 & \text{if } C \in [0,150], T \in [0,100], pH \in [1,14], t \in [1,60] \\ 0 & \text{otherwise} \end{cases}$$

Reward is given as $r_i = f_i (C_i, T_i, pH_i, t_i) - f_{i+1} (C_{i+1}, T_{i+1}, pH_{i+1}, t_{i+1})$ depending on how close the agent get to our targeted state. However, on the other hand, a punishment (or loss) would be addressed when the agent tries to go out of the normal range of corresponding parameters. The reward function is illustrated below:

$$r = \begin{cases} f(s) - f(s') & \\ -1 * x & x \in [1,4] \end{cases}$$

where *x* is the number of how many parameters are out of its normal range.

#### 3.1.2. *The proposed algorithms*

In this section, the decision support system based on RL algorithm is proposed to seek the optimal solution for minimizing manufacturing cost related to C, T, pH and t. Note that a random forest model is used to evaluate the difference between target and predicted result of the tested solution. The predictive error of it is limited to 1.5% in general. The pseudo code is demonstrated in Algorithm 1.

In the pseudo code of reinforcement main body, we need to give optimization targets to the system first (in terms of the values of k/s, L, a, b) with the basic parameters of the number of episodes E, the number of total steps N of agent in each episode, the learning rate $\alpha$ and the discount rate $\gamma$ (both in the range from 0 to 1, affecting the reward and future rewards obtaining from the states to update the Q(s, a) value). In the present case, E and N are given as 100 and 1000 respectively, while $\alpha$ and $\gamma$ are set as 0.05 and 0.8 respectively.

A Q-table then is created for the purples of recording state-action values Q(s, a), the columns of Q-table is in line with the actions in action space and the rows of it is in accordance with the limited states in state space. As mentioned, we know that the action space is $3^4$=81, and the state space in this case is composed by the solutions with four parameters $(C, T, pH, t)$ adjustable in the range of [0, 150], [0,100], [1, 14] and [1, 60] with the unit adjustment of 50, 10, 1 and 1 respectively, namely row number = 4 * 11 * 14 * 60 = 36960.

In the episode iteration, as E=100, agent would explore the environment 100 times from different initial state. The initialization of states in present algorithm is randomly choosing the four parameters ( $C, T, pH, t$ ) from corresponding range.

In each episode, the agent explores the environment from a random initial state, and interacts with the environment by observing next states as well as rewards. The agent chooses actions by using policy π (ε-greedy), this is a widely used policy in value based reinforcement learning and the pseudo code of it is illustrated in Algorithm 2.

---

**Algorithm 1: Reinforcement learning main body:**

**Input:** Targeted k/s, L, a, b value of the product and E, N, $\alpha$, $\gamma$
Create Q-table
**For** e < E do
   Initialize state vector $s_0$= (C, T, pH, t)
   **For** n < N **do**:
     Choose action from s using policy π (ε-greedy)
     Observe next state $s'$
     Estimate $f_i(s)$ and $f_{i+1}(s')$ to observe **r** ($r = f_i(s) - f_{i+1}(s')$)
     $Q^\pi(s, a) \leftarrow Q^\pi(s, a) + \alpha[r + \gamma max_{a'} Q^\pi(s', a') - Q^\pi(s, a)]$
     $s \leftarrow s'$
     n = n+1
   **End For**
n=0
e = e+1
**End For**

| **Algorithm 2:** Action choosing using policy π (ε-greedy) |
|---|
| **Input:** ε |
| **If** random() < ε |
|   randomly choose action from action space |
| **Else** |
|   choose action with maximum Q value. |
| **End if** |

ε-greedy policy helps the agent to find the best choice (maximum Q value) in present state to go to next state with a possibility of ε that may also randomly choose an action to get a next state randomly. This benefits the agent to explore the unexplored states without staying in exploitation of existing states in Q-table. In this study, ε was set as 0.88.

When action is selected, the Q value for that state-action pair (s, a) is updated on the basis of the reward received by the rule of $Q^\pi(s,a) \leftarrow Q^\pi(s,a) + \alpha[r + \gamma max_{a'} Q^\pi(s',a') - Q^\pi(s,a)]$. In particular, r in this update rule is calculated by means of the mentioned random forest regression model, according to different results of state s and next state s' ($r = f_i(s) - f_{i+1}(s')$ )

### 3.2. Results and discussion

In order to validate the applicability of this decision support system, a trail for finding solution in color fading ozonation to achieve target faded effect (targeted k/s, L, a, b = 0.8, 16, 21 and 71 respectively) from the original color (original k/s, L, a, b = 22.676, 64.97, 42.08 and 88.04 respectively) is investigated.

After 100 episodic explorations with 1000 steps in each episode, the result shows that [100, 60, 8, 31] is recommended to be used in the decision making of parameters $C, T, pH, t$ respectively in the color fading ozonation. The error tested by random forest predictive model is 3.211, which is a very good result with high acceptance of customer.

### 4. Conclusions

This paper proposed a reinforcement learning framework for decision support in textile manufacturing process. It described a decision making problem in textile manufacturing (color fading ozonation optimization) to the Markov Decision Process in terms of the tuple of {*S, A, P, R*}, and finding the optimization result by using Q-learning algorithm to train an agent to accumulate the Q(s, a) value by properly setting up the state, action, transition probability and reward function in this particular application. The result shows that this MDP model can well express the manufacturing problem and support the decision making.

**Acknowledgments**

The first author would like to express his gratitude to China Scholarship Council for supporting this study (CSC, Project NO. 201708420166).